\pdfoutput=1

\documentclass[11pt]{article}

\usepackage[preprint]{acl}

\usepackage{times}
\usepackage{latexsym}

\usepackage[T1]{fontenc}

\usepackage[utf8]{inputenc}

\usepackage{microtype}

\usepackage{inconsolata}

\usepackage{graphicx}

\usepackage{multirow}
\usepackage{todonotes}
\usepackage{subcaption}

\newcommand{\cola}{\textit{CoLA}}
\newcommand{\clinifact}{\textit{CliniFact}} 
\newcommand{\lcdbenchmark}{\textit{LCD bench (Summ)}} 

\title{Using tournaments to calculate AUROC for zero-shot classification with LLMs}

\author{
 \textbf{WonJin Yoon\textsuperscript{1,2}}, 
 \textbf{Ian Bulovic\textsuperscript{1}}, 
 \textbf{Timothy A. Miller\textsuperscript{1,2}} \\
\textsuperscript{1}Boston Children's Hospital \\
\textsuperscript{2}Harvard Medical School \\
 \small{
   \{firstname.lastname@childrens.harvard.edu\}
 }
}

\date{\today}

\begin{document}
\maketitle
\begin{abstract}
Large language models perform surprisingly well on many zero-shot classification tasks, but are difficult to fairly compare to supervised classifiers due to the lack of a modifiable decision boundary.
In this work, we propose and evaluate a method that transforms binary classification tasks into pairwise comparisons between instances within a dataset, using LLMs to produce relative rankings of those instances.
Repeated pairwise comparisons can be used to score instances using the Elo rating system (used in chess and other competitions), inducing a confidence ordering over instances in a dataset.
We evaluate scheduling algorithms for their ability to minimize comparisons, and show that our proposed algorithm leads to improved classification performance, while also providing more information than traditional zero-shot classification.

\end{abstract}


\section{Introduction}
In zero-shot classification with large language models (LLMs), a classification task is presented to the LLM in a prompt, and a discrete classification decision is extracted from the generated response.
These generated responses can be used to compute traditional scoring metrics of accuracy, precision, recall, and F1 score, to compare against other classification metrics.

Supervised methods, in contrast, typically output a probability distribution over possible outputs.
In the binary classification setting, real-world deployments of classifiers benefit from these explicit scores, as it allows downstream users to adjust decision thresholds to adapt classifiers to different use cases, with the same classifier potentially using different thresholds for different tasks.
In a \emph{screening} setting, recall is maximized at the expense of precision, because there may be such a high cost to missing cases that we can accept more false positives (e.g., doing research on out-of-hospital mortality, we want to capture as many potential cases as possible).
In a \emph{case-finding} setting, we may prefer to maximize precision at some expense to recall, for cases where we want our predicted cases to be a pure representation of that phenomenon (e.g., in a clinical trial recruitment setting, we want to err on the side of precision when it comes to satisfying inclusion and exclusion criteria).

Some commonly used instruments to evaluate a classifier at different decision points are the receiver-operator curve (ROC) and the precision-recall curve (PRC).
Measuring the area under the ROC and PRC curves (AUROC and AUPRC, respectively) can also be used as a single metric to summarize the amount of signal captured by a given classifier across all thresholds.

Zero-shot classification with LLMs does not have a straightforward way to perform the equivalent actions.
Unlike supervised methods, LLMs do not have the concept of a decision threshold, so they can only be evaluated at one point, and whether this point represents a bias towards high recall or high precision is not easy to measure.
For classification tasks where the class of interest has low prevalence, small changes in the number of positive predictions or in the decision threshold can have outsized effects on the F1 score. In such cases, it may be preferable to use threshold-independent metrics like AUROC, especially when comparing between classification paradigms (supervised vs. zero-shot LLMs).

In this paper, we describe a method for estimating AUROC or AUPRC for a zero-shot LLM by turning the classification task into a ranking task between pairs of instances.
We use the Elo rating system~\cite{elo1978rating}, used for chess and online videogames, to assign scores to instances in a dataset, updating them after ``wins'' or ``losses,'' and use the ordering this rating induces to compute AUROC.
Since this system requires more LLM calls than straightforward classification, we also investigate various tournament scheduling strategies and empirically measure which schedules minimize the number of competitions to reach competitive performance.
On three diverse datasets, \cola, \clinifact~and a patient outcome prediction task derived from \citealt{yoon2025lcd} (\lcdbenchmark), we find that our proposed method leads to stronger classification performance in terms of F1 scores.
Random scheduling performs surprisingly well, though the ``Swiss'' system used in chess and a novel graph-based method are also strong in most settings.


\section{Background and Related Work}
Our method is inspired by ratings systems like Elo, developed in chess~\cite{elo1978rating}, but applied to many other settings, including online video games.
This and other related systems (c.f., the Bradley-Terry model~\cite{bradley_rank_1952}) are implicitly grounded in probability distributions around player ratings, where differences in distributions can be used to calculate probability of one competitor winning against another.
These methods also consist of update methods where, after a competition between player $A$ and player $B$, respective Elo ratings can be updated based on the difference between scores.
For the Elo system, the updated rating for player A ($\hat{e}_A$) with Elo rating $e_A$ against player $B$ with Elo rating $e_B$ is:
\begin{equation}
\label{eqn:update}
\hat{e}_A = e_A + K * ([w_A] - P(e_A, e_B))
\end{equation}
where $[w_A]$ is $1$ if player $A$ wins or $0$ if not, and $K$ is a constant that modulates the rate of update.
We use $K=32$, as conventionally used in chess for players with few games played and lower ratings.
$P$ is a function representing the expected win probability of player $A$ against player $B$, defined as:
\begin{equation}
P(e_A, e_B) = {1 \over 1 + 10^{(e_B-e_A) \over 400}}
\end{equation}

Our method is motivated by the goal of generating ROC curves, and allowing practitioners to select appropriate cut-offs for downstream classification applications.
However, it shares a similarity with a recent thread of LLM-based research on the topic of estimating confidence of LLM answers.
If certainty could be estimated reliably, directly from LLM outputs, then ROC curves could be generated directly from those outputs.
However, recent work in this area shows mixed results, with some work showing difficulty finding correlations with certainty in zero-shot classification tasks~\cite{gao_position_2024}.
Other work does find some signal, but with white-box methods showing much greater signal than black-box methods~\cite{savage_large_2024}.
Work studying calibration claims that LLM uncertainties are calibrated for true/false question answering~\cite{kadavath_language_2022}, but that calibration only holds for the largest model they evaluate.


\section{Methods}

\subsection{Baseline}
To establish baseline performance, we use standard zero-shot prompting to get classification results and to measure the self-consistency of the LLMs. 
The prompt templates for our three datasets can be found in Appendix~\ref{app:prompts}.

Explicitly trading off precision and recall via zero-shot classification off generation alone is impossible, but we explored the possibility of ``prompt steering,'' that is, prompting to encourage the model to err on the side of precision or recall (see Appendix~\ref{app:prompts} for specific prompt steering variations).

\subsection{Tournament-based method}
Our proposed method is to induce an overall ranking of instances by converting our zero-shot prompts into pairwise comparison prompts, iteratively comparing pairs of instances from our dataset, and scoring each instance based on its ``wins'' and ``losses'' in the tournament.
These scores can be converted to probabilities, and the probabilities used as thresholds for decision-making, allowing for the computation of receiver-operator characteristic (ROC) or precision-recall (PR) curves.

\subsubsection{Tournament setup}
We randomly initialize every instance in our dataset $D$ to have an Elo rating distributed around 1000.
In each round $R$, we obtain a ranking of instances by Elo rating, and produce a set of pairs of matchups for that round via a scheduling algorithm.
The instance pair in each matchup is inserted into a pairwise comparison prompt template (see Appendix~\ref{app:prompts}), and an answer is extracted using regular expressions.\footnote{We use the Inspect Eval framework\cite{UK_AI_Safety_Institute_Inspect_AI_Framework_2024} from the UK AI Safety Institute.}
At the end of each round, Elo ratings for each instance are updated according to Equation~\ref{eqn:update}.
%


\begin{table*}[h!]
    \centering
    \resizebox{\textwidth}{!}{
    \begin{tabular}{|p{2.0cm}|c||c|c|c|c||c|c|c|c|}
        \hline
        \multirow{2}{=}{Model} & \multirow{2}{*}{Prompt style} & \multicolumn{4}{c||}{\cola} & \multicolumn{4}{c|}{\clinifact} \\
        \cline{3-10}
        & & Precision & Recall & F1 & Accuracy & Precision & Recall & F1 & Accuracy \\
        \hline
        \multirow{3}{=}{\raggedright Llama3.2-3b-Instruct} & Plain & 0.868 & 0.547 & 0.671 & 0.634 & 0.338 &    0.976 &    0.502 & 0.491 \\
        & Precision &0.759 & 0.458 & 0.571 & 0.533 & 0.312 & 0.928 & 0.467 & 0.443 \\
        & Recall & 0.754 & 0.660 & 0.704 & 0.619 & 0.324 & 0.976 & 0.487 & 0.459 \\
        \hline
        \multirow{3}{=}{GPT-4o-mini} & Plain & 0.914 & 0.877 & 0.895 & 0.858 & 0.660 & 0.747 & 0.701 & 0.832 \\
        & Precision & 0.910 & 0.890 & 0.900 & 0.863 & 0.670 & 0.735 & 0.701 & 0.835 \\
        & Recall & 0.869 & 0.929 & 0.898 & 0.854 & 0.660 & 0.747 & 0.701 & 0.832 \\
        \hline
    \end{tabular}
    }
    \caption{Performance of two LLMs on \cola~and \clinifact, when run in binary classification mode (Zero-shot, single run). Precision and recall prompts emphasize the greater cost of false positives or false negatives, respectively. Precision, recall, and F1 are defined in terms of the positive class, \emph{Acceptable} or \emph{True}, for our two datasets.}
    \label{tab:ml_results}
\end{table*}

\subsubsection{Scheduling strategies}
Since each instance is processed by the LLM one time in each round, a single round performs approximately as much computation as zero-shot classification.
However, after a single round the Elo scores and probabilities derived from them are unlikely to be very well sorted, so they likely require multiple rounds.
Therefore, we explore a variety of scheduling algorithms, with the goal of finding the algorithm that empirically minimizes the number of rounds required to obtain reliable AUROC values.
We evaluate the following scheduling strategies:

\textbf{Random}: In each round, we sample two instance indices at a time, without replacement, and match up those two indices in that round. There is no guarantee that two instances won't matchup repeatedly across rounds.

\textbf{Graph}: We model the set of instances as a graph with adjacency matrix $A$, initialized as the zero matrix. At the end of each round, we set $A_{i,j}=1$ for instance indices $i$ and $j$ that were matched up in that round. To schedule each round, we use the \emph{networkx} library~\cite{SciPyProceedings_11} to compute the minimum distance between all pairs of instances, and pair up instances that are furthest from each other first, again removing each instance from the pool for that round once it has been matched up. Instances that are not connected at all (all instances in the first round) are assigned distance equal to the size of the dataset. The motivation for this approach is the intuition that instances with long distance between them in the graph have the most uncertainty about their relative rankings and should be paired up. Compared to the \emph{Random} method, this will also reduce the number of repeated matches.

\textbf{Swiss}: This method, with empirically strong performance~\cite{sziklai_efficacy_2022}, ranks the instances by Elo rating, and splits the ordered list of instances into groups of size eight, and creates instance pairings within groups. In the variant we use, called ``The Dutch system,'' within each group (of size 8) player $i<4$ plays player $7-i$. A motivation for this approach is to allow players of similar rank to play each other. This may be beneficial in our case, where we might suspect local ordering to be more important for optimizing AUROC.


\begin{figure*}[h!]
    \centering
    \resizebox{\textwidth}{!}{
    \begin{subfigure}{0.45\textwidth}
        \centering
        \includegraphics[width=\textwidth]{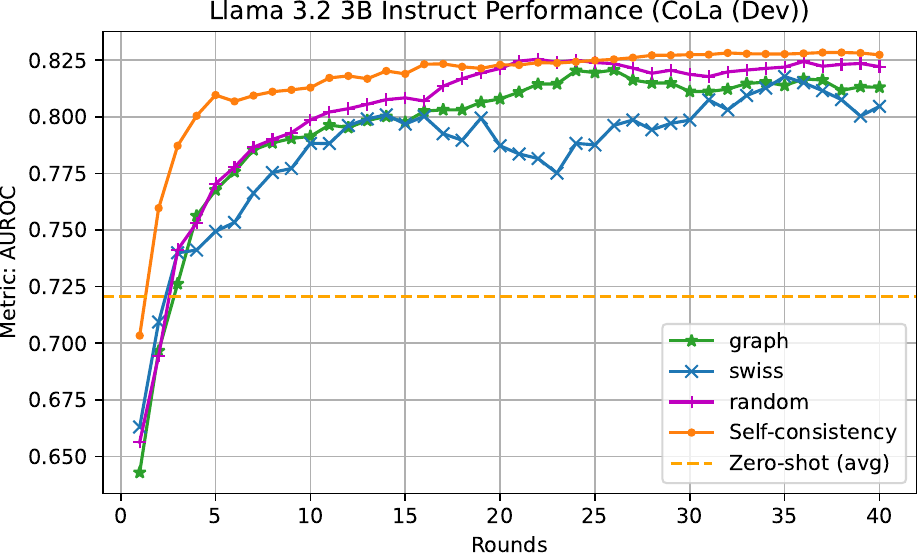}
        \caption{Llama-3.2-3B-Instruct performance on \cola}
    \end{subfigure}
    \hspace{5pt}
    \begin{subfigure}{0.45\textwidth}
        \centering
        \includegraphics[width=\textwidth]{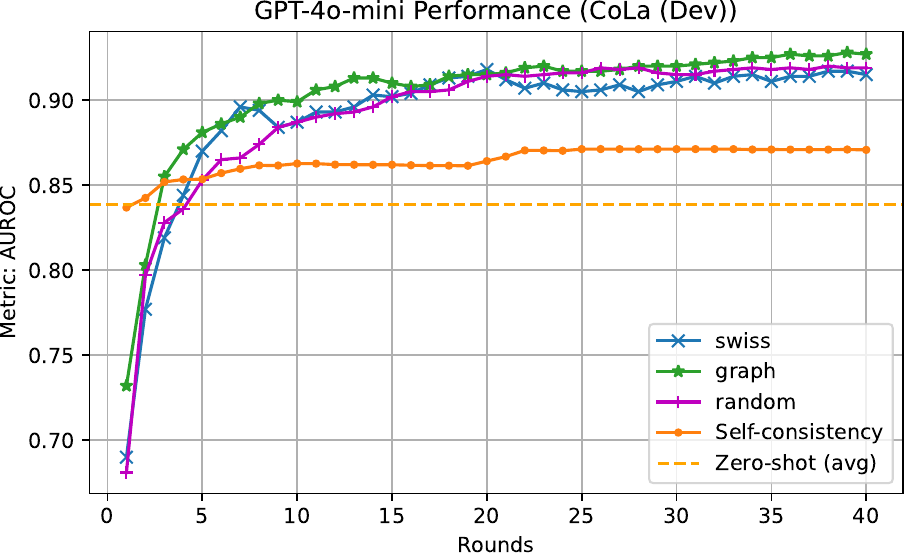}
        \caption{GPT-4o-mini performance on \cola}
        
    \end{subfigure}
    }
    \newline
    \\
    \resizebox{\textwidth}{!}{
    \begin{subfigure}{\columnwidth}
        \centering
        \includegraphics[width=\textwidth]{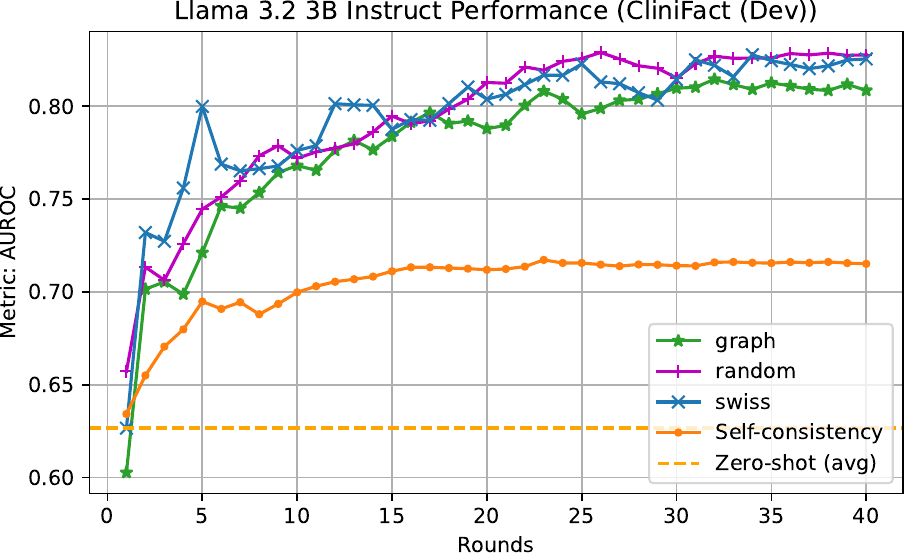}
        \caption{Llama-3.2-3B-Instruct performance on \clinifact}
    \end{subfigure}
    \hspace{5pt}
    \begin{subfigure}{\columnwidth}
        \centering
        \includegraphics[width=\textwidth]{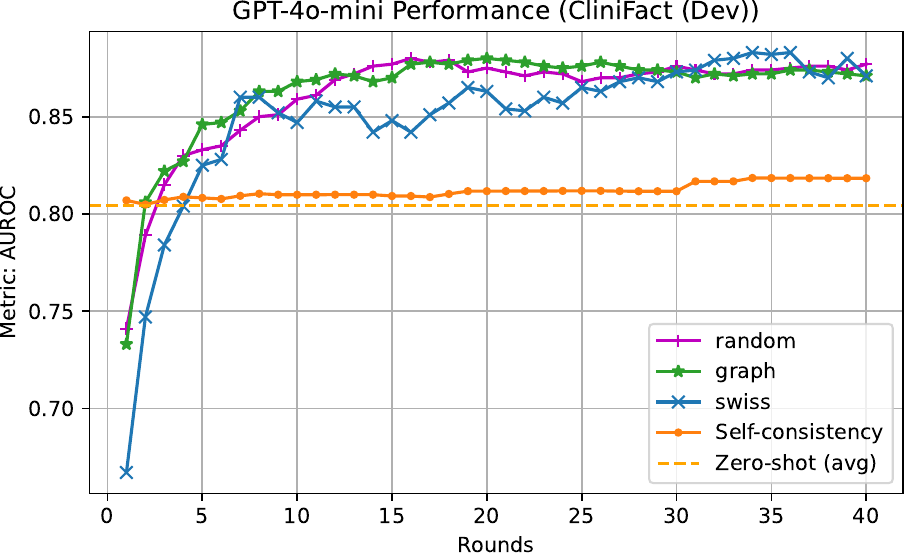}
        \caption{GPT-4o-mini performance on \clinifact}
    \end{subfigure}
    }
    \caption{Figures showing AUROC improvements across rounds for each dataset and model.}
    \label{fig:tiled_plots}
\end{figure*}
\begin{figure}[h!]
    \centering
    \setcounter{subfigure}{4}
    \begin{subfigure}{\columnwidth}
    \includegraphics[width=\textwidth]{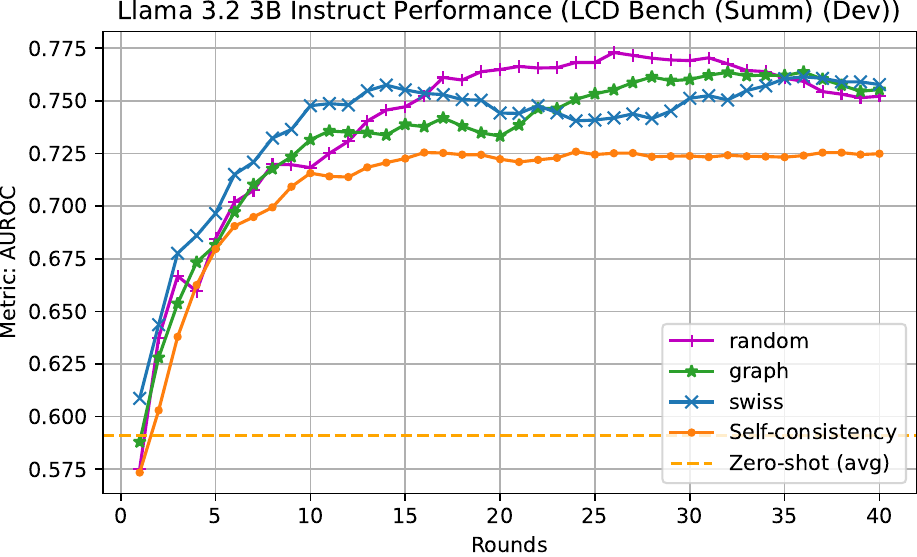}
    \end{subfigure}
    \caption{Figures showing \lcdbenchmark~results on Llama3.2-Instruct. Cloud-hosted models were not evaluated on this dataset in compliance with data usage restrictions.}
    \label{fig:tiled_plots2}
\end{figure}

\section{Evaluation}

We use Llama-3.2-3B-Instruct and GPT-4o-mini\footnote{Specifically, we use gpt-4o-mini-2024-07-18.} as representative models, ranging from smaller, more open models, to larger, more closed models. We further evaluated our approach on three different instruction-tuned models, Qwen 2.5 \cite{qwen2.5}, distilled DeepSeek R1 \cite{deepseekai2025deepseekr1}, and Phi-tiny-MoE \cite{abdin2024phi}, and present the results in the Appendix~\ref{sec:appendix-addional-exp}.
Our code to run these models uses the Inspect eval framework~\cite{UK_AI_Safety_Institute_Inspect_AI_Framework_2024}, which runs the Llama models using Huggingface transformers~\cite{wolf_transformers_2020}.
\subsection{Metrics}
We report precision, recall, F1, accuracy, and AUROC.

\paragraph{AUROC} AUROC is a threshold-independent metric that evaluates how well positive instances are ranked above negative ones, and it depends only on the relative ordering of the predicted scores.
Our proposed approach assigns an Elo rating system score to each instance in the dataset, which naturally induces a ranking of the instances and can serve as input to the metric when reversed.
For the self-consistency–based baseline, we iterate zero-shot prompt based classification with temperature of 1, which is the default value.  treat the number of positive answers ($n_{pos}$) across $N$ iterations as the confidence score ($C = n_{pos}/N$) for an instance, and use the confidence score to rank the instances to evaluate performance using the AUROC metric.
\subsection{Datasets}
To evaluate our proposed methods, we selected three datasets: one from the general domain (linguistics publications), one from the scientific domain (biomedical literature), and one from the clinical domain (electronic health records).
\paragraph{CoLA} The Corpus of Linguistic Acceptability (\cola)~\cite{warstadt2018neural} is a dataset consisting of sentences from linguistics publications, labeled with whether they were considered \emph{acceptable} or not.
We use version 1.1 of \cola, and evaluate on the in-domain development set.

\paragraph{CliniFact} Our second data source is \clinifact~\cite{zhang_dataset_2025}, a dataset of scientific claims from clinical trial protocols labeled for whether they are supported or not by corresponding publications.

\paragraph{LCD Benchmark (Summary)}
The Long Clinical Document Benchmark (LCD Benchmark) \cite{yoon2025lcd} is designed to evaluate language models on 30-day out-of-hospital mortality prediction using patient discharge notes from the MIMIC-IV \cite{johnson2023mimic} database. Because the original notes often exceed 8,000 tokens, we created summarized versions to reduce input length, enabling us to process pairs of samples plus prompts without exceeding the GPU memory available in a typical local research environment. Unless otherwise specified, we use the term \textbf{\lcdbenchmark}~in this paper to refer to this summarized version of the dataset. In compliance with MIMIC/PhysioNet data use restrictions, we did not run this dataset on GPT.\footnote{https://physionet.org/news/post/gpt-responsible-use}

\subsection{Results}
\paragraph{Preliminary} Table~\ref{tab:ml_results} shows the result of the zero-shot classifiers on our two datasets.
GPT-4o-mini significantly outperforms Llama-3.2-3b at both tasks, as might be expected.
F1 scores are respectable for \cola, but show imbalanced precision and recall.
For Clinifact, Llama-3.2-3b struggles significantly, but GPT-4o-mini performs respectably, with higher overall F1 and better balance between precision and recall.
On all configurations, prompt-based steering towards higher \emph{precision} is completely ineffective, while steering towards higher \emph{recall} does succeed for \cola.
This preliminary experiment highlights the need for methods that can be steered toward precision or recall.

Figure~\ref{fig:tiled_plots} and ~\ref{fig:tiled_plots2} shows AUROC values across tournament rounds for \cola,~\clinifact~and~\lcdbenchmark. Different schedulers, and two baselines (zero-shot prompting and self-consistency) are represented as color lines. For the self-consistency baseline, the x-axis (``Rounds'') represents the number of iterations ($N$). The zero-shot prompting baseline is averaged over 40 independent runs and shown as horizontal dashed lines.

Across the datasets and LLMs we experiment with, our tournament approach consistently outperforms the two baselines, regardless of the scheduler selection. One exception is the \cola~dataset, where the self-consistency baseline occasionally performs better. We observe comparable results for the other LLMs (Figure~\ref{fig:appendix_tiled_plots_cola},~\ref{fig:appendix_tiled_plots_clinifact}, and~\ref{fig:appendix_tiled_plots_lcd}) in Appendix~\ref{sec:appendix-addional-exp}.
The Random scheduler is strong, obtaining the high overall AUROC for all the datasets in the early rounds, but the differences between the schedulers converge over the rounds. 
Table~\ref{tab:tourney_results} shows the F1 scores that are obtained by finding the optimal cutoff point on the ROC curve for the optimal AUROC round for each tournament experiment.
%
While the added benefit of AUROC scores and thresholds mean it would be sufficient to match the performance in the zero-shot setting, in fact we find that the tournament-style method leads to greatly improved F1 scores.
For \cola, Llama improves from 0.671 to 0.866, and GPT-4o-mini improves from 0.895 to 0.910.
For \clinifact, F1 scores with Llama improve from 0.502 to 0.651, and with GPT-4o-mini they improve from 0.701 to 0.753.


\begin{table}[h!]
    \centering
    \resizebox{\columnwidth}{!}{
    \begin{tabular}{|p{2.0cm}||c|c|c||c|c|c|}
        \hline
        \multirow{2}{=}{Model} & \multicolumn{3}{c||}{\cola} & \multicolumn{3}{c|}{\clinifact} \\
        \cline{2-7}
        & Acc & F1 & AUC  & Acc & F1 & AUC \\
        \hline
        Llama3.2-3b-Instruct &   0.801 & 0.866 & 0.825 & 0.816 & 0.651 & 0.828 \\
        \hline
        GPT-4o-mini          & 0.873 & 0.910 & 0.928 & 0.873 & 0.753 & 0.883 \\
        \hline
    \end{tabular}
    }
    \caption{Optimal performance of two LLMs at \cola~and\clinifact, when run in tournament mode. Precision, recall, and F1 are defined in terms of the positive class, \emph{Acceptable} or \emph{True}, for our two datasets. The \emph{Graph} scheduler was used for the GPT-4o-mini \cola{} results, while the \emph{Random} scheduler was used for the Llama \cola{} results. For \clinifact, \emph{Random} and \emph{Swiss} scheduler were used for Llama and GPT, respectively.}
    \label{tab:tourney_results}
\end{table}


\section{Discussion and Conclusion}
One major benefit of this approach relative to other methods for estimating uncertainty is that it can be computed from generations only.
While closed models like ChatGPT make log probabilities available in their API, there is no guarantee that such functionality will always exist.
Our method only relies on the core functionality of LLMs, the ability to generate text.

A drawback of our approach is its requirement of multiple passes over the dataset.
Our results show asymptotic behavior after 10-20 rounds, but the smaller dataset does asymptote quickly, raising the possibility that the optimal number of rounds is a function of dataset size.
Future work should explore this relationship to judge the feasibility of applying this method to larger datasets.

The code implementing our methods and generating these results is available at: \\ \url{https://github.com/Machine-Learning-for-Medical-Language/cnlp_llm}.

\section*{Acknowledgments}
Research reported in this publication was supported by the National Institute Of Mental Health and National Library of Medicine of the National Institutes of Health
under Award Numbers R01MH126977, R01LM012973, and R01LM012976. The content is solely the responsibility of the authors and does not
necessarily represent the official views of the National Institutes of Health.
\section*{Limitations}
This work evaluates on three datasets, chosen to have different properties, but still limiting the generalizeability of the work.
Our selection of the models, which, though representative, cannot represent all models.
While our method applies in black box scenarios that other methods do not (like looking at first token probability), we did not include a comprehensive suite of alternative methods for comparison of performance.
The method we describe also requires more inference time than simple classification.
While we test scheduling algorithms to improve this performance, it will always require the equivalent of multiple inference passes per instance.


\bibliography{custom}

\clearpage
\appendix

\section{Prompt Templates}
\label{app:prompts}
\subsection{Single instance prompts}
For the \cola~dataset, our baseline (single instance classification) prompt was: 
\begin{quote}
Please read the following sentence and decide whether it is "acceptable" in a linguistic sense (i.e., grammatical). Don't explain your reasoning, just answer "Yes" (acceptable) or "No" (unacceptable) on a new line. \{Input sentence\}
    \end{quote}

For the \clinifact~dataset, our baseline (single instance classification) prompt was: 
\begin{quote}
Instruction: Given a scientific claim and an abstract, determine if the abstract reports positive results (TRUE) or not (FALSE) about the claim. The task is to classify the pair claim abstract as follows: TRUE: if the abstract provides positive support for the claim. FALSE: if the abstract provides negative or inconclusive support for the claim or if the abstract provides contextual or background information without directly reporting results about the claim. \{Input sentence\}    
\end{quote}

For the \lcdbenchmark~dataset, our baseline (single instance classification) prompt was: 
\begin{quote}
Instruction: You are an experienced critical care physician with expertise in accurately predicting patient outcomes. Predict whether a patient will die within 30 days after discharge (TRUE) or survive (FALSE), based on the provided discharge summary note. If an abstract presents inconclusive findings or does not provide information relevant to the claim, the answer is FALSE. Don't explain your reasoning, just answer TRUE or FALSE on a new line. \{Input sentence\}    
\end{quote}

\subsection{Precision and recall-enhanced prompts}
We attempted to guide the model towards higher precision or recall with variations of the prompt stressing the consequences of false positives or negatives. These prompts are the same as the baseline prompt, with the following text added.

For the version favoring precision: 
\begin{quote}
The consequences for wrongly guessing \{Positive class\} are worse than the consequences for wrongly guessing \{Negative class\}.
\end{quote}

For the version favoring recall: 
\begin{quote}
The consequences for wrongly guessing \{Negative class\} are worse than the consequences for wrongly guessing \{Positive class\}.
\end{quote}

\subsection{Pairwise comparison prompts}
For the \cola\ dataset, the pairwise prompt was: 

\begin{quote}
You are an expert linguist deciding whether sentences are grammatically acceptable or not. Your task is to take in a pair of sentences and decide which is more acceptable. The output format should be \{"choice": <Sentence>, "reasoning": <your reasoning>\}, where <Sentence> should be the more acceptable or less unacceptable sentence, either "Sentence 1" or "Sentence 2". Here are the two sentences.

Sentence 1: \{text1\}

Sentence 2: \{text2\}
\end{quote}

For the \clinifact\ dataset, the pairwise prompt was:

\begin{quote}
    You are a biomedical researcher evaluating whether given scientific claims and their corresponding abstracts report positive results (TRUE) or not (FALSE). If an abstract presents inconclusive findings or does not provide information relevant to the claim, the answer is FALSE. Your task is to compare two claim-abstract pairs and determine which one is more likely to be classified as TRUE. If both pairs should be answered "TRUE", choose the one with higher confidence. The output format should be \{"choice": <Pair>, "reasoning": <your reasoning>\}, where <Pair> should be the claim abstract pair more likely to have the answer TRUE, either "Pair 1" or "Pair 2". Here are the two questions.
    
    Pair 1: \{text1\}
    
    Pair 2: \{text2\}
\end{quote}

\section{Additional experiments on various LLMs}\label{sec:appendix-addional-exp}

\begin{figure}[h!]
    \centering
   
    \begin{subfigure}{0.45\textwidth}
        \centering
        \includegraphics[width=\textwidth]{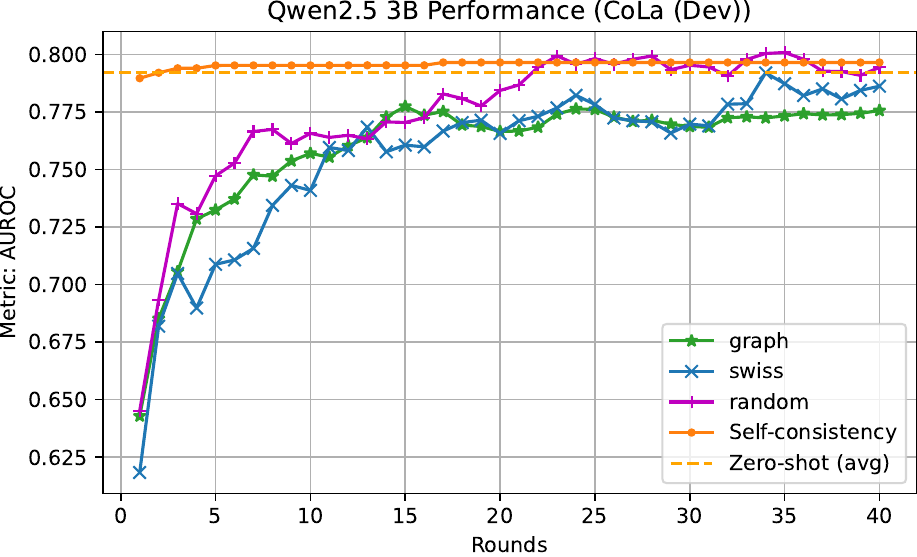}
        \caption{Qwen2.5-3B Instruct performance on \cola}
    \end{subfigure}

    \vspace{1em}
    
    \begin{subfigure}{0.45\textwidth}
        \centering
        \includegraphics[width=\textwidth]{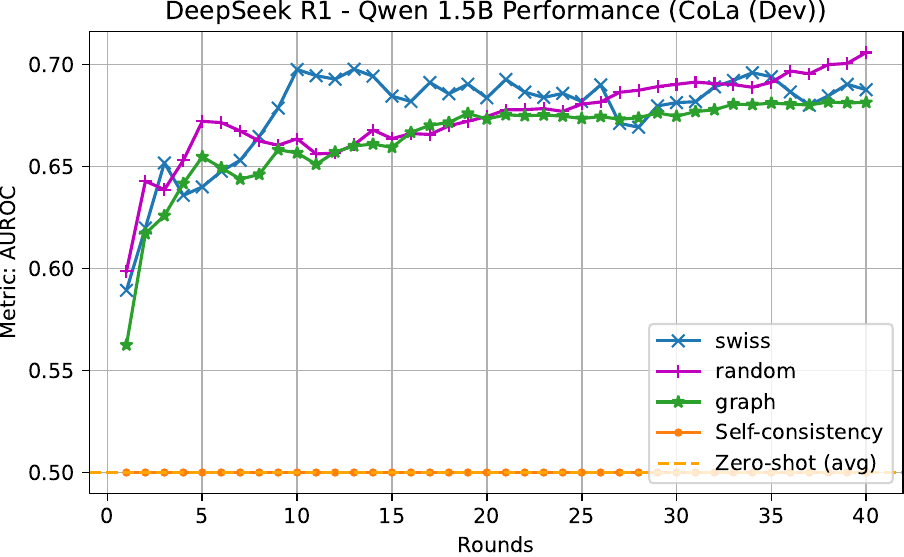}
        \caption{DeepSeek-R1-Distill-Qwen-1.5B performance on \cola}
        
    \end{subfigure}

    \vspace{1em}
    
    \begin{subfigure}{0.45\textwidth}
        \centering
        \includegraphics[width=\textwidth]{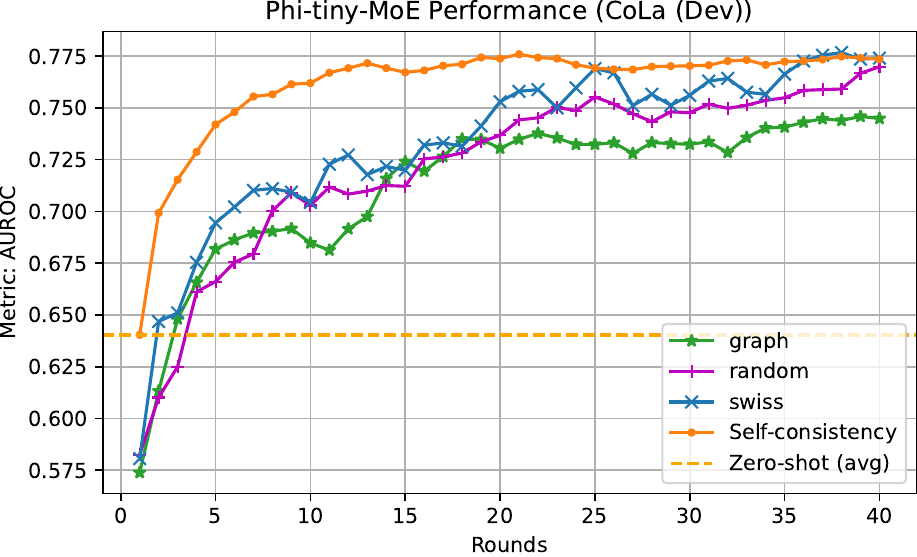}
        \caption{Phi-tiny-MoE-instruct performance on \cola}
        
    \end{subfigure}

    \caption{Figures showing AUROC improvements across rounds for \cola.}
    \label{fig:appendix_tiled_plots_cola}
\end{figure}

\begin{figure}[h!]
    \centering
   
    \begin{subfigure}{0.45\textwidth}
        \centering
        \includegraphics[width=\textwidth]{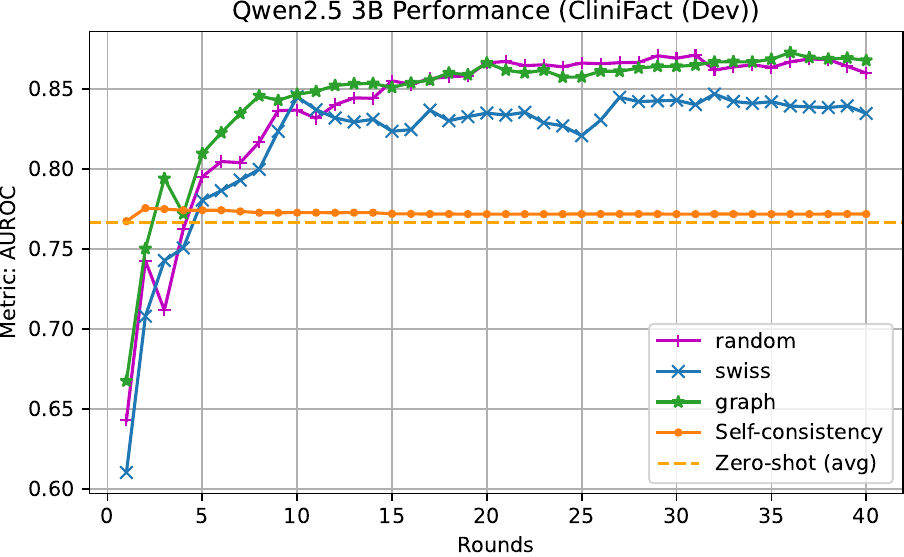}
        \caption{Qwen2.5-3B Instruct performance on \clinifact}
    \end{subfigure}

    \vspace{1em}
    
    \begin{subfigure}{0.45\textwidth}
        \centering
        \includegraphics[width=\textwidth]{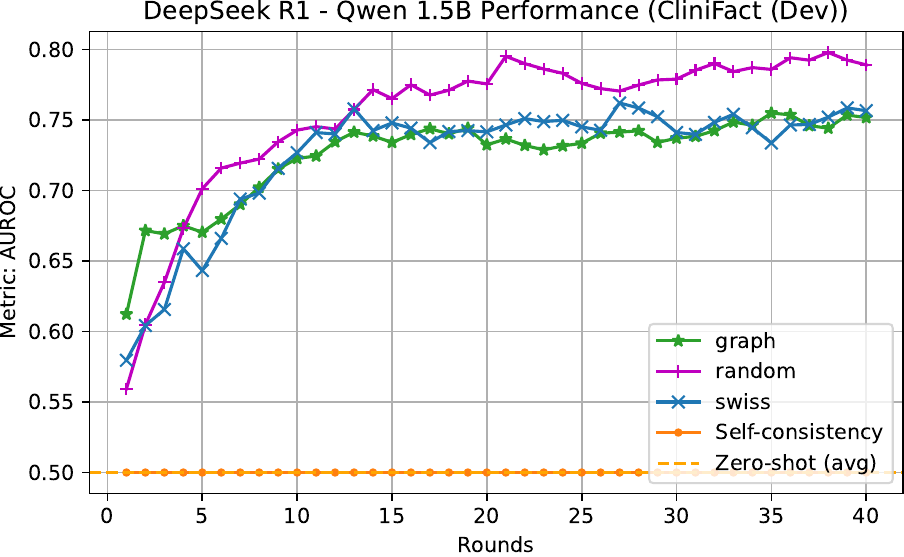}
        \caption{DeepSeek-R1-Distill-Qwen-1.5B performance on \clinifact}
        
    \end{subfigure}

    \vspace{1em}
    
    \begin{subfigure}{0.45\textwidth}
        \centering
        \includegraphics[width=\textwidth]{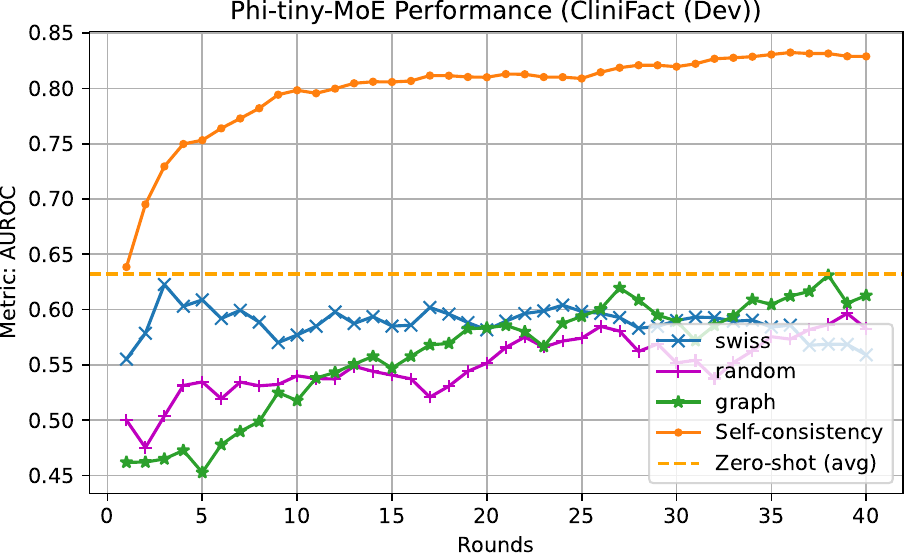}
        \caption{Phi-tiny-MoE-instruct performance on \clinifact}
        
    \end{subfigure}

    \caption{Figures showing AUROC improvements across rounds for \clinifact.}
    \label{fig:appendix_tiled_plots_clinifact}
\end{figure}

\begin{figure}[h!]
    \centering
   
    \begin{subfigure}{0.45\textwidth}
        \centering
        \includegraphics[width=\textwidth]{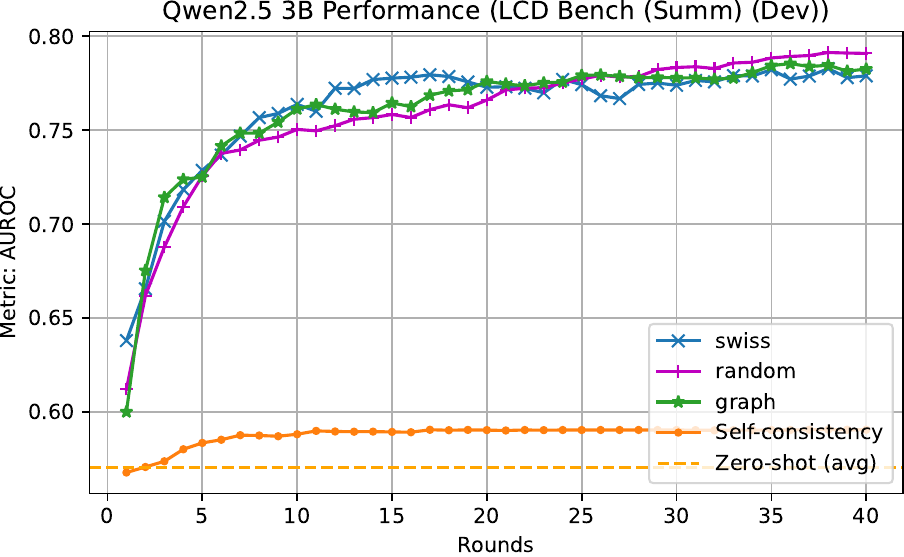}
        \caption{Qwen2.5-3B Instruct performance on \lcdbenchmark}
    \end{subfigure}

    \vspace{1em}
    
    \begin{subfigure}{0.45\textwidth}
        \centering
        \includegraphics[width=\textwidth]{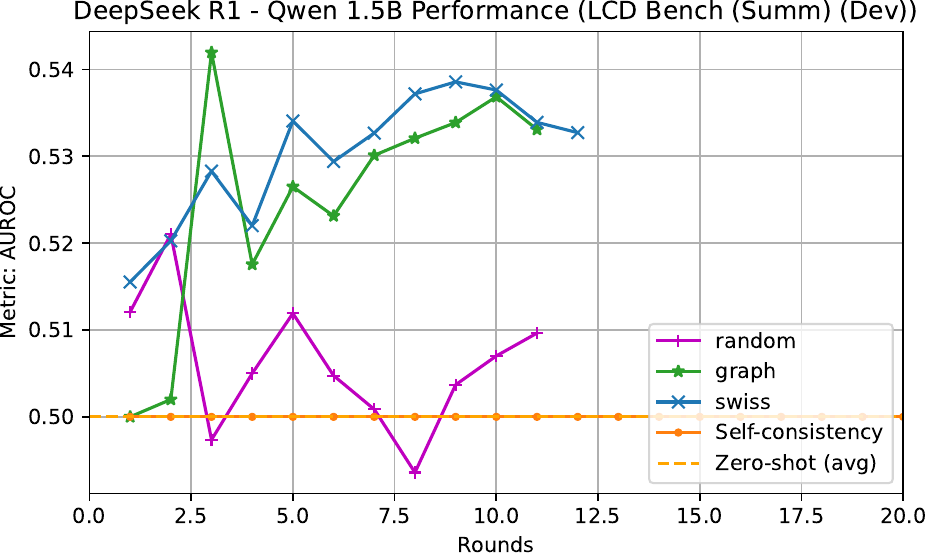}
        \caption{DeepSeek-R1-Distill-Qwen-1.5B performance on \lcdbenchmark}
        
    \end{subfigure}

    \vspace{1em}
    
    \begin{subfigure}{0.45\textwidth}
        \centering
        \includegraphics[width=\textwidth]{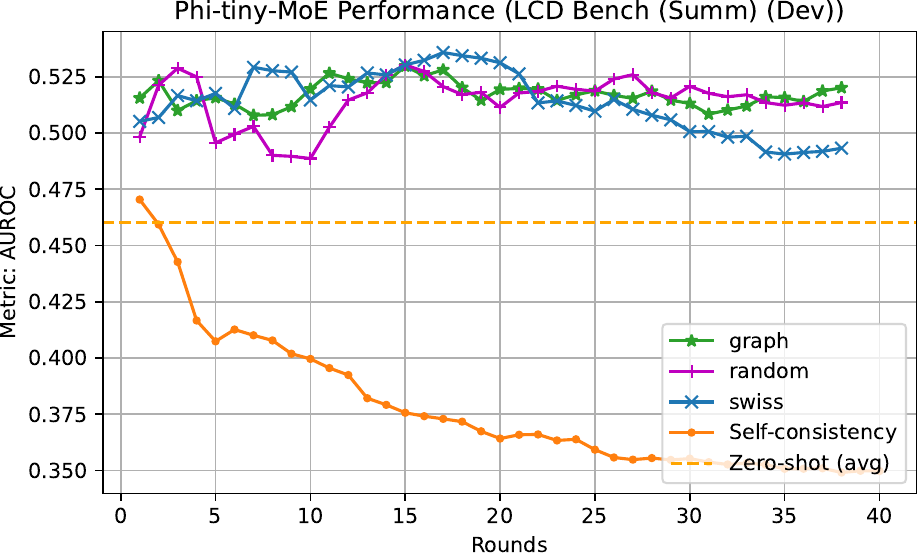}
        \caption{Phi-tiny-MoE-instruct performance on \lcdbenchmark}
        
    \end{subfigure}

    \caption{Figures showing AUROC improvements across rounds for \lcdbenchmark.}
    \label{fig:appendix_tiled_plots_lcd}
\end{figure}

Figure~\ref{fig:appendix_tiled_plots_cola},~\ref{fig:appendix_tiled_plots_clinifact}, and~\ref{fig:appendix_tiled_plots_lcd} present additional experimental results from our method on different models and datasets. Since AUROC was chosen as the evaluation metric, a score of 0.5 represents the baseline of a random or uninformative classifier, and a score below 0.5 indicates that the model is making predictions in the opposite direction of the correct answers.

An interesting result can be observed with the DeepSeek-R1-Distill-Qwen-1.5B model. This model did not follow the instruction in the binary classification task prompts to respond with [``Yes'' or ``No''] or [``True'' or ``False''] and instead produced other types of sentences (e.g.,``Okay, so I'm trying ...''), which led to all its answers being marked incorrect. One possible interpretation is that, as this model was primarily developed for reasoning tasks, it may be inclined to output such sentences. This remains an interesting topic for future investigation.

Another noteworthy result comes from the Phi-tiny-MoE model on the LCD bench (Figure~\ref{fig:appendix_tiled_plots_lcd} (c)), which achieved performance below 0.5. One possible explanation is that the model may have interpreted the task as predicting survival rather than mortality. This is also a point worth further exploration in future work.

Some experiments in Figure~\ref{fig:appendix_tiled_plots_lcd} are not completed within the given time window due to limited GPU resources.

\clearpage

\end{document}